\documentclass{article}
\usepackage{spconf,amsmath,graphicx,hyperref}
\usepackage{algorithm}
\usepackage{algorithmic}
\usepackage{multirow} 
\usepackage{booktabs}
\usepackage{amsmath}
\usepackage{amsthm}
\usepackage{amssymb}

\usepackage{xcolor}
\title{Training-Free Test-Time Adaptation with Brownian Distance Covariance in Vision-Language Models}
\name{Yi Zhang$^1$, Chun-Wun Cheng$^2$, Angelica~I.~Aviles-Rivero$^3$, Zhihai He$^4$, Liang-Jie Zhang$^1$\sthanks{Corresponding author.}}
\address{
\small{$^1$College of Computer Science and Software Engineering, Shenzhen University, China \quad  $^2$University of Cambridge, Cambridge, UK}\\
\small{$^3$Yau Mathematical Sciences Center, Tsinghua University, Beijing, China \quad $^4$Southern University of Science and Technology, Shenzhen, China}
}
\begin{document}
%\ninept
%
\maketitle
\begin{abstract}
Vision-language models suffer performance degradation under domain shift, limiting real-world applicability. Existing test-time adaptation methods are computationally intensive, rely on back-propagation, and often focus on single modalities.
To address these issues, we propose \textit{Training-free Test-Time Adaptation with Brownian Distance Covariance} (TaTa). TaTa leverages Brownian Distance Covariance—a powerful statistical measure that captures both linear and nonlinear dependencies via pairwise distances—to dynamically adapt VLMs to new domains without training or back-propagation. This not only improves efficiency but also enhances stability by avoiding disruptive weight updates.
TaTa further integrates attribute-enhanced prompting to improve vision-language inference with descriptive visual cues. Combined with dynamic clustering and pseudo-label refinement, it effectively recalibrates the model for novel visual contexts. Experiments across diverse datasets show that TaTa significantly reduces computational cost while achieving state-of-the-art performance in domain and cross-dataset generalization.
\end{abstract}
\section{Introduction}

\label{sec:intro}
Vision-language models (VLMs) such as  Align~\cite{jia2021scaling} and CLIP~\cite{radford2021clip} learn aligned multimodal representations from large-scale image-text data, enabling effective zero-shot inference by comparing image and text embeddings in a shared space~\cite{lu2019vilbert, kim2021vilt, desai2021virtex}. However, their performance degrades under significant domain or distribution shifts. Test-time adaptation (TTA) addresses this by adapting models using only unlabeled test data, without accessing source data or modifying training~\cite{shu2022test,chen2022contrastive}. This approach aligns well with practical scenarios where only the trained model is available, without access to the source data or authorization to alter the original training procedure. In such cases, models need to quickly adapt to new tasks. 

\begin{figure}[t!]
\begin{center}
\centerline{\includegraphics[width=1\linewidth]{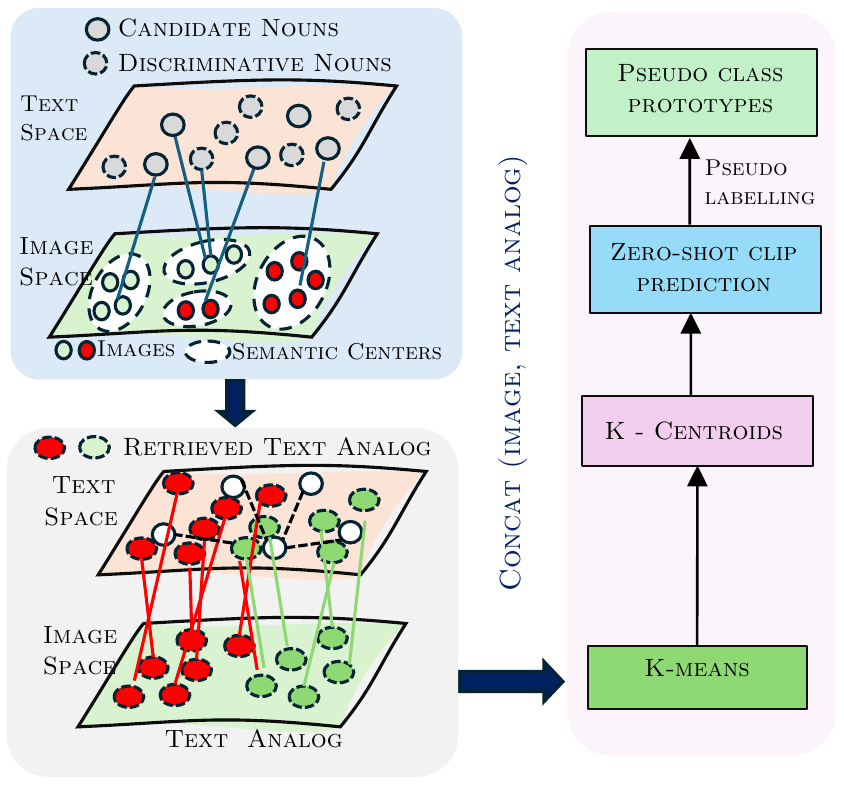}}
\vspace{-15pt}
\caption{
The process involves using Pseudo-Labeling with Multimodal-Assisted Clustering to categorize nouns into semantic centers and create a discriminative text space. Nouns for each image are matched in this space, and by concatenating image and text features, K-means clustering forms centroids and clusters. 
}

\label{fig:pl}
\end{center}
\vspace{-25pt}
\end{figure}

Recent prompt-based TTA methods such as TPT~\cite{shu2022test} and DiffTPT~\cite{feng2023diverse} improve VLM adaptation but require computationally expensive optimization~\cite{yoonc}. 
TDA~\cite{karmanov2024efficient} avoids backpropagation and offers a training-free alternative, but has two key limitations: 1) TDA uses cosine similarity to measure the similarity between test image features and those in caches. This method only considers marginal distributions and linear relationships, potentially missing complex, non-linear dependencies in high-dimensional features. 2) TDA neglects the importance of visual attributes in prompts, relying solely on class labels or simple descriptions. This oversight limits the ability to capture the full semantic richness of images, resulting in less accurate and nuanced classifications, especially in complex or ambiguous scenarios.

To address these issues, we propose Training-free TTA method with Brownian Distance Covariance for VLMs. Brownian distance covariance can capture more comprehensive statistics by considering joint distribution. It effectively models both linear and non-linear relationships between query and support images by measuring the discrepancy between the joint distribution of their features and the product of their marginals.
The final prediction in our proposed TaTa is a combination of vision-vision and vision-language inference. For vision–vision inference, we use K-means with multimodal clues to cluster test images and assign pseudo-labels, as shown in Figure~\ref{fig:pl}. BDC measures similarity between test features and pseudo-labeled centroids, capturing both linear and non-linear dependencies. 
For vision–language inference, we enrich prompts with visual attributes to improve semantic alignment.
Finally, to address the prediction bias, we propose a soft-voting strategy that incorporates knowledge from nearest neighbor samples.

In summary, our contribution can be summarized as: 1) We present a streamlined and robust method for adapting vision-language models during testing without requiring additional training. 2) We introduce pseudo labeling with a dynamic clustering with multimodality, and adopt brownian distance covariance as the metrics for vision-vision inference, and we exploit visual attributes description words to enhance the prompting for vision-language inference. To mitigate pseudo-labels bias, we introduce a soft-voting strategy. 3) Extensive experimental results show that TaTa significantly surpasses existing state-of-the-art methods in both domain generalization and cross-dataset generalization tasks.

\section{Method}
\label{sec:method}

\begin{figure*}[t]
\begin{center}
\centerline{\includegraphics[width=1\linewidth]{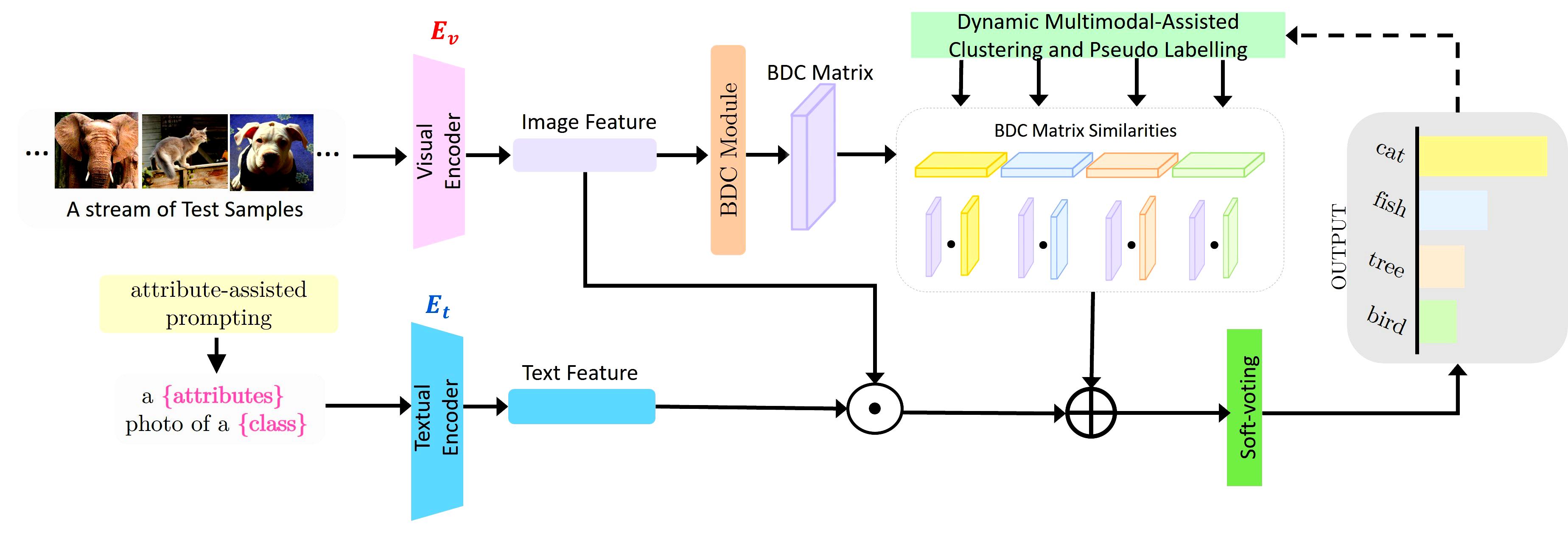}}
\vspace{-18pt}
\caption{An overview of our proposed TaTa. The final inference in TaTa combines Vision-Vision and Vision-Language Inference. 
}
\label{fig:overview}

\end{center}
\vspace{-18pt}
\end{figure*}

\textbf{Method Overview. }
Figure~\ref{fig:overview} illustrates our proposed TaTa framework for test-time adaptation, which enhances prediction by leveraging knowledge from the test stream and adapting image features. The final inference combines Vision–Vision(V-V) and Vision–Language(V-L) pathways.
For a test image, we extract visual features using $E_v$ and combine them with textual analogs to form a multimodal feature $f_m$. In Vision–Vision Inference, we compute Brownian Distance Covariance (BDC) between $f_m$ and pseudo-labeled class prototypes derived from clustering. Simultaneously, for Vision–Language Inference, we construct attribute-enhanced prompts and compute cosine similarity between image features and corresponding text embeddings.
A dynamic dictionary $\mathcal{D}$ is maintained, mapping class names to clusters. Correctly classified test samples update $\mathcal{D}$ with their multimodal features, continuously refining class centroids and reducing pseudo-label bias. A soft-voting mechanism incorporating nearest-neighbor knowledge further improves prediction robustness.

\subsection{Dynamic Pseudo-Labeling with Multimodal-Assisted Clustering}
A dynamic dictionary $\mathcal{D}$ stores class labels as keys and corresponding clusters as values. When a test image is correctly classified, its multimodal feature (concatenated image and text representations) is incorporated into the corresponding class in $\mathcal{D}$, updating the clusters and centroids. 

\textbf{Dynamic Multimodal-Assisted Clustering:}
To construct a discriminative textual space without known test class names, we use a subset of WordNet nouns~\cite{WordNet}. As shown in Figure~\ref{fig:pl}. For a dataset with $N$ classes, visual features extracted by $E_v$ are first clustered via K-Means to obtain initial centroids $C = \{C_i\}_{i=1}^N$. Using CLIP, we assign each WordNet noun to the closest semantic center based on similarity: $p(y=i \mid \mathbf{T_k}) = \frac{\exp(\operatorname{sim}(f_{t_k}, C_i))}{\sum_{j=1}^N \exp(\operatorname{sim}(f_{t_k}, C_j))},$
where $\mathbf{T_k}$ is a text prompt such as “a photo of {noun}” and $f_{t_k}$ is encoded by $E_t$. Top-$k_1$ nouns per center are selected to form a representative textual set $\{\bar{T}_m\}_{m=1}^H$ where $H = N \times k_1$ ($k_1=5$ empirically).

For each image feature $f_{v_h}$ of image $x_h$, we compute its textual analog $f_{t_h}$ by aggregating noun embeddings $\{\bar{f}_{t_h}\}_{h=1}^H$ with similarity-weighted coefficients:
$f_{t_h} = \sum_{j=1}^{H} p(\bar{f}_{t_j} \mid f_{v_h}) \bar{f}_{t_j},$
$p(\bar{f}_{t_j} \mid f_{v_h}) = \frac{\exp(\operatorname{sim}(f_{v_h}, \bar{f}_{t_j}) / \tilde{\tau})}{\sum_{l=1}^H \exp(\operatorname{sim}(f_{v_h}, \bar{f}_{t_l}) / \tilde{\tau})},$
where $\tilde{\tau}=0.005$ (following TAC~\cite{li2023image}). Finally, K-Means is applied to the concatenated features $[f_{t_d}, f_{v_d}]_{d=1}^D$ to obtain refined centroids $\bar{C} = \{\bar{C}_i\}_{i=1}^N$.

\textbf{Pseudo Labeling: } 
Text features are generated from class-based prompts using $E_t$. Pseudo-labels are assigned to each centroid $\bar{C}_i$ via zero-shot similarity comparison:
$p(y=m \mid \bar{C}_i) = \frac{\exp(\operatorname{sim}(f_{t_m}, \bar{C}_i)/\tau)}{\sum_{j=1}^N \exp(\operatorname{sim}(f_{t_j}, \bar{C}_i)/\tau)},$
where $\tau$ is the softmax temperature. The resulting multimodal prototypes $\bar{C}$ serve as class representatives for inference.

\subsection{BDC for V-V Inference}
We design a Brownian Distance Covariance (BDC) module $\mathcal{B}(x)$ to capture both linear and nonlinear dependencies between feature representations. BDC measures dependence via distance covariance, based on pairwise distances between sampless~\cite{szekely2009brownian,DeepBDC-CVPR2022}.
Given two random vectors $\mathbf{X}$ and $\mathbf{Y}$, the BDC is computed as follows:
1. Compute Euclidean distance matrices:
   $a_{ij} = \|\mathbf{X}_i - \mathbf{X}_j\|_2$, $b_{ij} = \|\mathbf{Y}_i - \mathbf{Y}_j\|_2$
2. Center the distance matrices:
   $\bar{A}_{ij} = a_{ij} - \frac{1}{n}\sum_k a_{ik} - \frac{1}{n}\sum_l a_{lj} + \frac{1}{n^2}\sum_{k,l} a_{kl}$
   (similarly for $\bar{B}_{ij}$)
3. Calculate the distance covariance:
   $\text{dCov}^2(\mathbf{X}, \mathbf{Y}) = \frac{1}{n^2} \sum_{i,j} \bar{A}_{ij} \bar{B}_{ij}$.
BDC is non-zero only if features are dependent, capturing both linear and nonlinear relationships. It is parameter-free and training-free.

Inference: 
For each pseudo-labeled centroid $\bar{C}_i$ from clustering, we compute its BDC matrix $\mathcal{P}_{bdc}^i = \mathcal{B}(\bar{C}_i)$. For a test image $x_h$ with multimodal feature $f_m = [f_{v_h}, f_{v_t}]$, we compute its BDC matrix $f_{bdc}^m = \mathcal{B}(f_m)$. The prediction probability is:
\[
p_{vv}(y=i|x_h)= \frac{\exp\left(\text{dCov}^2 (f_{bdc}^m, \mathcal{P}_{bdc}^i)\right)}{\sum_{j=1}^{N} \exp\left(\text{dCov}^2 (f_{bdc}^m, \mathcal{P}_{bdc}^j)\right)}.
\]

\subsection{Attribute-assisted prompting for V-L Inference} Including descriptions of visual context can make the text more accurate, thereby improving the consistency between visual and linguistic modalities. This leads to better performance in CLIP's zero-shot inference. For example, ``a photo of a grey koala on the tree" is more accurate than ``a photo of a cat". Therefore, for the vision-language inference, we aim to exploit visual attributes to assist in prompting. First, we build a list $\Pi_t \triangleq \{\pi_i\}_{i=1}^{k_2}$ which contains $k_2 = 2000$ texts that describe the common visual attributes. The list $\Pi_t$ is derived from previous works~\cite{zhang2024concept,zhang2024conceptual} and includes texts describing materials, patterns, colors, interactions, backgrounds, and more. Following CLIP's zero-shot setup, we concatenate each $\pi_i$ with a hand-crafted prompt $\psi$ ``\texttt{The photo is} ..." to create a attribute-specific textual input $\{\psi; \pi_i\}$. This combined input is then fed into the text encoder $E_t$ to generate the attribute textual features $A_{t} \triangleq \{a_t^i\}_{i=1}^{k_2}$. For the image $x_{te}$, we use the image encoder $E_v$ to extract the image feature $f_v$, then we compute the cosine similarity with each attribute textual feature. We choose the attribute texts in $\Pi_t$ whose text features have the top-$k_2$ largest similarity score, to form the text ``a \{attributes\} photo of a \{class\}". We then extract its feature $f_t$ using $E_t$ and perform the vision-language inference as:
\begin{equation}
\label{eq-P_vl}
    p_{vl}(y=i|x_{te})= \frac{\exp(\operatorname{sim}(f_v, f_t^i/\tau)}{\sum_{j=1}^N \exp(\operatorname{sim}(f_v, f_t^j)/\tau)},
\end{equation}
where $\tau$ is a temperature coefficient and ``sim" denotes cosine similarity. $p_{vl}(y=i|x_{te})$ represents the prediction probability for $x_{te}$ belonging to label $i$.

\subsection{Inference Fusion}
We then combine $p_{vv}$ and $p_{vl}$ to get the final prediction, 
\begin{equation}
\label{eq-LOGIT}
{p(y=i|x_{te})} = \alpha p_{vv}(y=i|x_{te}) + p_{vl}(y=i|x_{te})
\end{equation}
Prediction refinement is achieved by leveraging the knowledge from nearest neighbor samples, based on the premise that semantically similar images lie close in the feature space and are likely to share the same label. 
For a target image \(x_{te}\), we extract its feature \(f_v = E_v(x_{te})\), which is used to retrieve neighboring samples. The prediction for \(x_{te}\) is then refined by aggregating the class probabilities of these neighbors via soft voting:
$\hat{p}(y=i \mid x_{te}) = \frac{1}{k_3 + 1} \left( \sum_{j=1}^{k_3} p'_j + p(y=i \mid x_{te}) \right)$
where \(k_3\) is the number of neighbors and \(\hat{p}(y=i \mid x_{te})\) is the refined probability for class \(i\). 

\vspace{-5pt}

%%%%%%%%%%%%%%%%%%%%%%%%%%

\renewcommand\arraystretch{0.95}
\begin{table*}[ht]
  \caption{\textbf{Comparison with state-of-the-art methods on the Cross-Domain Benchmark.}
  % We evaluated TaTa against several baselines, including CLIP, train-time adaptation methods (CoOp and CoCoOp), and test-time adaptation methods (TPT, DiffTPT, and TDA). Train-time methods are trained on ImageNet and then tested directly on these 10 datasets.
  }
  \centering
  \resizebox{\linewidth}{!}{
    \begin{tabular}{l*{11}c}
      \toprule
      Method & Aircraft & Caltech101 & Cars & DTD & EuroSAT & Flower102 & Food101 & Pets & SUN397 & UCF101 & Average \\
      \midrule
      CLIP-ViT-B/16 & 23.22 & 93.55 & 66.11 & 45.04 & 50.42 & 66.99 & 82.86 & 86.92 & 65.63 & 65.16 & 64.59 \\
      \midrule
      CoOp & 18.47 & 93.70 & 64.51 & 41.92 & 46.39 & 68.71 & 85.30 & 89.14 & 64.15 & 66.55 & 63.88 \\
      CoCoOp & 22.29 & 93.79 & 64.90 & 45.45 & 39.23 & 70.85 & 83.97 & \textbf{90.46} & 66.89 & 68.44 & 64.63 \\
      \midrule
      TPT & 24.78 & \underline{94.16} & 66.87 & \underline{47.75} & 42.44 & 68.98 & 84.67 & 87.79 & 65.50 & 68.04 & 65.10 \\
      DiffTPT & \underline{25.60} & 92.49 & 67.01 & 47.00 & 43.13 & 70.10 & \underline{87.23} & 88.22 & 65.74 & 62.67 &65.47 \\
      TDA & 23.91 & \textbf{94.24} & \underline{67.28} & 47.40 & \underline{58.00} & \textbf{71.42} & 86.14 & 88.63 & \underline{67.62} & \underline{70.66} & \underline{67.53} \\
      \textbf{TaTa (Ours)} & \textbf{26.68} & 93.82 & \textbf{68.04} & \textbf{49.57} & \textbf{59.67} & \underline{71.28} & \textbf{89.45} & \underline{89.91} & \textbf{68.64} & \textbf{73.54} & \textbf{69.06} \\
      \bottomrule
    \end{tabular}
  } 
  % \caption{\textbf{Comparison with state-of-the-art methods on the Cross-Domain Benchmark.} We evaluated TaTa against several baselines, including CLIP, train-time adaptation methods (CoOp and CoCoOp), and test-time adaptation methods (TPT, DiffTPT, and TDA). Train-time methods are trained on ImageNet and then tested directly on these 10 datasets.
  % }
  \label{tab:fine-grained}
  \vspace{-15pt}
\end{table*}

%%%%%%%%%%%%%%%%%%%%%%%%%%%
  
\renewcommand\arraystretch{0.9}
\begin{table}[t]
\caption{\textbf{Comparison on domain generalization tasks}.
% We compare our methods with baselines on ViT-B/16 CLIP. The baselines include CLIP, train-time adaptation methods (CoOp, CoCoOp, and Tip-Adapter), and test-time
% adaptation methods (TPT, DiffTPT, and TDA). Train-time methods are trained on ImageNet and then tested directly on other datasets.
% The metric 'Average' represents the mean accuracy across all five datasets, while 'OOD Average' is calculated by the four OOD datasets.
% The results of all the baselines are from the TDA paper~\cite{karmanov2024efficient}.
}
  \centering
  \resizebox{\linewidth}{!}{
  \begin{tabular}{l*{6}c}
    \toprule
    {Method}      & ImageNet   &  -A  &  -V2  & -R 
    & -S   & {OOD Average}     \\
%   \midrule
%   CLIP-ResNet-50       &59.81&	23.24&	52.91	&{60.72}	&35.48&	46.43&	43.09           \\ 
%   \midrule
%   CoOp            &  \textbf{63.33} &  23.06  &   55.40     &  56.60   &  34.67   &     46.61   &     42.43       \\
  
%   CoCoOp         &  \underline{62.81} &  23.32  &   55.72     &  57.74  &  34.48   &     46.81   &    42.82   \\
% Tip-Adapter & 62.03 & 23.13 & 53.97 & 60.35  & 35.74 & 47.04 & 43.30 \\
%   \midrule
%   TPT          &  60.74 & 26.67  &    54.70   &  59.11   &  35.09   &    47.26    &     43.89    \\
%   DiffTPT & 60.80 & \underline{31.06} & \underline{55.80} & 58.80 & 37.10 & 48.71 & 45.69 \\
%     TDA  & 61.35 & {30.29} & 55.54
%     & \underline{62.58} & \underline{38.12} & \underline{49.58} & \underline{46.63} \\
%     \textbf{TaTa (Ours)} & 62.17 & \textbf{31.81} & \textbf{56.96}
%     & \textbf{64.22} & \textbf{39.45} & \textbf{50.92} & \textbf{48.11} \\
%   \midrule
  \midrule
  CLIP      & 68.34&	49.89&	61.88&	{77.65}&	48.24&	59.42             \\
  \midrule
  CoOp          &  \textbf{71.51} &  49.71  &   64.20     &  75.21   &  47.99       &   59.28  \\
  
  CoCoOp         &  \underline{71.02}  & 50.63  & 64.07       &  76.18   &  48.75       &   59.91  \\
 Tip-Adapter& 70.75 & 51.04 & 63.41 & 77.76 & 48.88  & 60.27 \\
  \midrule
  TPT           &  68.98  & 54.77  & 63.45       &  77.06   &  47.94        &   60.81 \\
  DiffTPT           &  70.30  & 55.68  & 65.10       &  75.00   &  46.80        &   60.52 \\
  TDA  & 69.51 & \underline{60.11} & 64.67 & \underline{80.24} & \underline{50.54 } &\underline{63.89} \\
  \textbf{TaTa (Ours)} & 70.63 & \textbf{61.87} & \textbf{65.37} & \textbf{81.78} & \textbf{52.39}  & \textbf{65.28} \\
    \bottomrule
  \end{tabular}}
% \caption{\textbf{Comparison with other methods on domain generalization benchmark}. We compare our methods with baselines on ViT-B/16 CLIP. The baselines include CLIP, train-time adaptation methods (CoOp, CoCoOp, and Tip-Adapter), and test-time
% adaptation methods (TPT, DiffTPT, and TDA). Train-time methods are trained on ImageNet and then tested directly on other datasets.
% The metric 'Average' represents the mean accuracy across all five datasets, while 'OOD Average' is calculated by the four OOD datasets excluding ImageNet.
% % The results of all the baselines are from the TDA paper~\cite{karmanov2024efficient}.
% }
\vspace{-10pt}
\label{tab:ood-main}
\end{table}
  
\renewcommand\arraystretch{1.0}
\begin{table}[h]
\caption{\textbf{Comparison of testing time and accuracy.}  }
\centering
\begin{tabular}{lccccc}
\toprule
    Method & Testing Time &  Accuracy &Gain\\ \midrule
    CLIP-ViT-B/16 &12min &68.34 &0\\
    TPT &12h\ 50min & 68.98  &{+0.64}\\
    TDA & {16min} & {69.51} & {+1.17}\\
    \textbf{TaTa (Ours)} & \textbf{13.5min} & \textbf{70.63}\ &\textbf{+2.29}\\
\bottomrule
\end{tabular}
% \caption{ Comparison of testing time and accuracy. }

\label{tab:time}
\vspace{-10pt}
\end{table}

\section{Experiments}
\label{sec:experiments}
\subsection{Benchmark Settings}

\noindent\textbf{Baselines. }In this paper, we comprehensively compare our proposed TaTa with seven state-of-the-art methods designed for vision-language models, including CLIP \cite{radford2021clip}, CoOp \cite{zhou2022learning}, CoCoOp \cite{zhou2022conditional}, Tip-Adapter \cite{zhang2022tip}, TPT \cite{shu2022test}, DiffTPT \cite{feng2023diverse}, and TDA \cite{karmanov2024efficient}. Among these, CoOp, CoCoOp, and Tip-Adapter are train-time adaptation methods, while TPT, DiffTPT, and TDA are test-time adaptation methods. Train-time adaptation methods are trained on a 16-shot ImageNet set~\cite{imagenet_cvpr09} and tested on other datasets, while test-time adaptation methods are fine-tuned directly on target datasets using the test set, without utilizing the ImageNet training set.

\noindent\textbf{Implementation details.}
\label{sec:impl_details}
Our method utilizes ViT-B/16 CLIP for our experiments. Following prior methods~\cite{radford2021clip}, we adhere to data preprocessing protocols in CLIP. We empirically set $k_1 = 5$, $k_3 = 4$, and $\alpha = 1.75$. 
Following~\cite{karmanov2024efficient,shu2022test}, we report the top-1 accuracy (\%), a standard classification criterion, as our evaluation metric.
% All experiments were conducted using a single NVIDIA RTX3090 GPU.

\subsection{Performance Analysis}
\noindent\textbf{Domain Generalization. }In Table~\ref{tab:ood-main}, we display the experimental results of the domain generalization benchmark, comparing our proposed TaTa to various baselines. 
% To comprehensively evaluate the effectiveness of our method, we compare it with the base model CLIP, train-time adaptation methods, and test-time adaptation methods. 
Compared to CLIP, TaTa shows significant improvements, with performance gains of up to 5.86\%  for OOD average.
Notably, since train-time methods (CoOp and CoCoOp) are trained on the labeled training set of ImageNet, it is expected that they perform better on ImageNet. Interestingly, our method even outperforms Tip-Adapter on ImageNet, despite it being training-free but having access to the labeled training set. For the OOD average, TaTa consistently and substantially outperforms the train-time methods. 
When compared to test-time adaptation methods, our method outperforms all others by a large margin for both dataset-specific and OOD average results, except for ImageNet-V2. TaTa surpasses TDA by  +1.39\% on OOD average. These results demonstrate the effectiveness of our method and its strong test-time adaptation capability.

\noindent\textbf{Cross-dataset Generalization}
Table~\ref{tab:fine-grained} shows the comparison with state-of-the-art methods on the cross-dataset generalization task. It is obvious that our proposed TaTa substantially outperforms other methods on average. Compared to the second-best method TDA, our method achieves a performance gain of 1.53\% . In particular, TaTa outperforms TDA by up to 2.77\% on Aircraft and 2.88\% on UCF101. Compared to train-time adaptation methods, TaTa surpasses them by up to 5.18\% on average. These results highlight the strong test-time adaptation capability of our method across diverse class datasets.
This feature is particularly beneficial for vision-language models like CLIP, as it allows them to classify a wide range of classes in image classification without requiring additional training.

\begin{table}[t]
\caption{\textbf{Effectiveness of different components in TaTa.} AAP refers to Attribute-Assisted Prompting. BDC represents the BDC module. MAC stands for Multimodal Assisted Clustering. SV denotes Soft-voting. The last row is our TaTa.}
\centering
\resizebox{\linewidth}{!}{
\begin{tabular}{lcc}
\toprule
Method & ImageNet   & OOD Average \\ \midrule
CLIP                    & 68.34 & 59.42  \\
\, + AAP               & 69.05 & 61.21  \\
\, + AAP + BDC          & 70.03   & 64.15    \\
\, + AAP + BDC +MAC         & 70.42   & 64.85    \\
\, \textbf{+ AAP + BDC + MAC  + SV}   & \textbf{70.63} & \textbf{65.28} \\
\bottomrule
\vspace{-15pt}
\end{tabular}
}

\label{table:components}
\end{table}
\vspace{-5pt}
\subsection{Ablation study}
\vspace{-5pt}

\noindent\textbf{Contributions of major algorithm components.}
We conduct experiments on ViT-B/16 CLIP. As shown in Table~\ref{table:components}, all three components significantly contribute to the overall performance improvement. 
Among them, the BDC module provides the largest performance gain, yielding a 2.94\% improvement in the OOD average. This demonstrates that BDC provides an efficient metric for classification. 

\noindent\textbf{Efficiency Comparison.} The comprehensive results are reported in Table~\ref{tab:time}. As shown, our method achieves new state-of-the-art performance in less time, exhibiting remarkable efficiency in test-time adaptation for VLMs. Compared to CLIP, TaTa gains 2.29\% more accuracy with just 1.5 additional minutes.
% Notably TPT takes 12 hours and 50 minutes, yet show inferior performance. In contrast, our method demonstrates strong efficiency and effectiveness. 
Compared to TDA, TaTa outperforms it by 1.12\% and reduces testing time by 1.5 minutes.

\section{Conclusion}
\label{sec:conclusion}

We introduce TaTa, a training-free test-time adaptation method that addresses domain shift in vision-language models. By utilizing Brownian Distance Covariance and avoiding backpropagation, it reduces computational cost while maintaining stability. Through dynamic multimodal clustering and pseudo-labeling, TaTa enhances generalization performance across tasks. Experimental results confirm its effectiveness and practical potential, establishing TaTa as a robust solution for deploying adaptive vision-language models.

\subsubsection*{Acknowledgments}
CWC and AIAR acknowledge support from the Swiss National Science Foundation (SNSF) under grant number 20HW-1 220785.  
AIAR gratefully acknowledges the support of the Yau Mathematical Sciences Center, Tsinghua University. This work is also supported by the Tsinghua University Dushi Program.

\bibliographystyle{IEEEbib}
\bibliography{refs}

\end{document}